\documentclass[11pt,a4paper]{article}
\usepackage{times,latexsym}
\usepackage{url}
\usepackage[T1]{fontenc}
\usepackage{booktabs}
\usepackage{tabularx}
\usepackage[dvipsnames, table]{xcolor}
\usepackage{array}
\usepackage{amsmath}
\usepackage{amssymb}
\usepackage{bbm}
\usepackage{graphicx}
\usepackage{caption}
\usepackage{subcaption}
\usepackage{siunitx}
\usepackage{multirow}
\usepackage{url}
\usepackage{microtype}
\usepackage{makecell}
\usepackage[english]{babel}

\usepackage[acceptedWithA]{tacl2021v1}
\usepackage{xspace,mfirstuc,tabulary}

\newif\iftaclinstructions
\taclinstructionsfalse % AUTHORS: do NOT set this to true
\iftaclinstructions
\renewcommand{\confidential}{}
\renewcommand{\anonsubtext}{(No author info supplied here, for consistency with
TACL-submission anonymization requirements)}
\newcommand{\instr}
\fi

\iftaclpubformat % this "if" is set by the choice of options

\else

\fi

\newcolumntype{R}{>{\raggedleft\arraybackslash}X}

\newcommand{\beluga}{\textsc{BeLUGA}\xspace}

\newcommand{\fscore}{$F_1$\xspace}
\newcommand{\g}[1][]{\ensuremath{\text{G}_{\text{#1}}}\xspace}
\newcommand{\gfunc}{\g[func]}
\newcommand{\gclass}{\g[class]}
\newcommand{\gtype}{\g[type]}
\newcommand{\gseen}{\g[seen]}
\newcommand{\gstandard}{\g[standard]}
\newcommand{\p}{\ensuremath{p}\xspace}
\newcommand{\x}{\ensuremath{\mathbf{x}}\xspace}
\newcommand{\X}{\ensuremath{\mathcal{X}}\xspace}
\newcommand{\y}{\ensuremath{y}\xspace}
\newcommand{\Y}{\ensuremath{\mathcal{Y}}\xspace}

\newcommand{\T}{\ensuremath{\mathcal{T}}\xspace}
\newcommand{\C}{\ensuremath{\mathcal{C}}\xspace}
\newcommand{\Ttrain}{\ensuremath{\mathcal{T}_\text{train}}\xspace}
\newcommand{\Tval}{\ensuremath{\mathcal{T}_\text{val}}\xspace}
\newcommand{\Ttest}{\ensuremath{\mathcal{T}_\text{test}}\xspace}

\newcommand{\loss}{\ensuremath{\ell}\xspace}

\newcommand{\nfunc}{\ensuremath{n_\text{func}}\xspace}

\newcommand{\nclass}{\ensuremath{n_\text{class}}\xspace}
\newcommand{\D}{\ensuremath{\mathcal{D}}\xspace}
\newcommand{\F}{\ensuremath{\mathcal{F}}\xspace}

\newcommand{\phenom}{\ensuremath{\mathcal{P}}\xspace}
\newcommand{\Dtrain}{\ensuremath{\mathcal{D}_\text{train}}\xspace}
\newcommand{\Dval}{\ensuremath{\mathcal{D}_\text{val}}\xspace}
\newcommand{\Dtest}{\ensuremath{\mathcal{D}_\text{test}}\xspace}
\newcommand{\R}{\ensuremath{\mathcal{R}}\xspace}
\newcommand{\iidAndT}{IID$\rightarrow$T\xspace}
\newcommand{\iidPlusT}{IID$+$T\xspace}
\newcommand{\iidAndIidPlusT}{IID$\rightarrow$(IID$+$T)\xspace}
\DeclareMathOperator*{\argmin}{argmin}
\DeclareMathOperator*{\argmax}{argmax}
\makeatletter
\@namedef{Changes@AuthorColor}{red}
\colorlet{Changes@Color}{red}
\makeatother
\title{Cross-functional Analysis of Generalisation in Behavioural Learning}

\author{Pedro Henrique Luz de Araujo$^{1,2}$ \and
  Benjamin Roth$^{1,3}$ \\
  $^1$Faculty of Computer Science, University of Vienna, Vienna, Austria \\
  $^2$UniVie Doctoral School Computer Science, Vienna, Austria\\
  $^3$Faculty of Philological and Cultural Studies, University of Vienna, Vienna, Austria\\
 \texttt{\{pedro.henrique.luz.de.araujo, benjamin.roth\}@univie.ac.at} \\ 
  }

\date{}

\begin{document}
\maketitle
\begin{abstract}
%Behavioural testing of natural language processing models has been proposed as an evaluation methodology to address the shortcomings of computing metrics on a held-out i.i.d. test set.
In behavioural testing, system functionalities underrepresented in the standard evaluation setting (with a held-out test set) are validated through controlled input-output pairs.
% that highlight important phenomena that would be underrepresented in the standard evaluation setting of computing metrics on a held-out i.i.d. test set.
% that align with human intuitions about how to perform the examined task.
Optimising performance on the behavioural tests during training (\emph{behavioural learning}) would improve coverage of phenomena not sufficiently represented in the i.i.d.\ data and could lead to seemingly more robust models.
However, there is the risk that the model narrowly captures spurious correlations from the behavioural test suite, leading to overestimation and misrepresentation of model performance---one of the original pitfalls of traditional evaluation.\\
In this work, we introduce \beluga, an analysis method for evaluating behavioural learning considering generalisation across dimensions of different granularity levels.
We optimise behaviour-specific loss functions and evaluate models on several partitions of the behavioural test suite controlled to leave out specific phenomena.
An aggregate score measures generalisation to unseen functionalities (or overfitting).
We use \beluga to examine three representative NLP tasks (sentiment analysis, paraphrase identification and reading comprehension) and compare the impact of a diverse set of regularisation and domain generalisation methods on generalisation performance.\footnote{Our code is available on \url{https://github.com/peluz/beluga}.}
%Our evaluation scheme both makes explicit the desired system traits and allows quantifying overfitting to them.
\end{abstract}

\section{Introduction}

% Motivate problem
% WHat is behavioral testing
The standard paradigm for evaluating natural language processing (NLP) models is to compute correctness metrics on a held-out test set from the same distribution as the training set \cite{linzen_accelerateHLLG_ACL_2020}.
If the test set is large and diverse, this may be a good measure of average performance, but it fails to account for the worst-case performance \cite{Sagawa2020Distributionally}.
By exploiting correlations in the training data, models work well in most cases but fail in those where the correlations do not hold \cite{niven_kao_probeNLC_ACL_2019,mccoy_etal_right4wrongreasons_ACL_20219,zellers_etal_hellaSwag_ACL_2019}, leading to overestimation of model performance in the wild \cite{ribeiro_etal_checklist_ACL_2020}.
Furthermore, standard evaluation does not indicate the sources of model failure \cite{wu_etal_errudite_acl_2019} and disregards important model properties such as fairness \cite{ma_etal_dyaboard_neurips_2021}.

Behavioural testing \cite{rottger_etal_hateCheck_ACL_2021,ribeiro_etal_checklist_ACL_2020} has been proposed as a complementary evaluation framework, where model capabilities are systematically validated by examining its responses to specific stimuli.
This is done through test suites composed of input-output pairs where the input addresses specific linguistic or social phenomena and the output is the expected behaviour given the input.
The suites can be seen as controlled challenge datasets \cite{belinkov_glass_analNLP_TACL_2019} aligned with human intuitions about how the agent should perform the task \cite{linzen_accelerateHLLG_ACL_2020}.

% Behavioural test taxonomy
In this work, we understand test suites as a hierarchy of functionality classes, functionalities, and test cases \cite{rottger_etal_hateCheck_ACL_2021}.
\emph{Functionality classes} stand at the highest level, capturing system capabilities like fairness, robustness and negation.
They are composed of \emph{functionalities} that target finer-grained facets of the capability.
For example, a test suite for sentiment analysis can include the functionality ``negation of positive statement should be negative'' inside the Negation class.
Finally, each functionality is composed of \emph{test cases}, the input-output pairs used to validate model behaviour.
For the functionality above, an example test case could be the input ``The movie was not good'' and the expected output ``negative'', under the assumption that the non-negated sentence is positive.

Though behavioural test suites identify model weaknesses, the question of what to do with such feedback is not trivial. 
While test suite creators argue that these tools can aid the development of better models \cite{rottger_etal_hateCheck_ACL_2021} and lead to improvements in the tested tasks \cite{ribeiro_etal_checklist_ACL_2020}, how to act on the feedback concretely is not discussed.
% \cite{tu2020empirical} show that models can generalise from minority patterns in the training data.

One common approach is fine-tuning on data targeting the failure cases, which previous work has shown can improve performance in these same cases \cite{malon2022fast,liu_etal_InocFineTune_NACACL_2019,mccoy_etal_right4wrongreasons_ACL_20219}.
But this practice overlooks the possibility of models overfitting to the covered tests and consequently overestimates model performance.
Even if one takes care to split the behavioural test cases into disjoint sets for training and testing, models can still leverage data artifacts such as word-label co-occurrences to achieve seemingly good performance that is over-optimistic and does not align with out-of-distribution (OOD) performance. 

This creates the following dilemma: either one does not use the feedback from test suites for model development and loses the chance to improve model trustworthiness;
or one uses it to address model shortcomings (e.g.\ by training on similar data)---and run the risk of overfitting to the covered cases.
Prior work \cite{luzdearaujo2022checking,rozen2019diversify} has addressed this in part by employing structured cross-validation, where a model is trained and evaluated on different sets of phenomena.
However, the analyses have been so far restricted to limited settings where only one task, training configuration and test type is examined.
Moreover, these studies have not examined how different regularisation and generalisation mechanisms influence generalisation.
% Furthermore, they do not discuss how to compare different systems considering both performance on suite and general i.i.d.\ data and no investigation on the impact of regularisation and generalisation algorithms.

In this paper, we introduce \beluga, a general method for \textit{Be}havioural \textit{L}earning \textit{U}nified \textit{G}eneralisation \textit{A}nalysis.
By training and evaluating on several partitions of test suite and i.i.d.\ data, we measure model performance on unseen phenomena, such as held-out functionality and functionality classes.
This structured cross-validation approach yields scores that better characterise model performance on uncovered behavioural tests than the ones obtained by over-optimistic i.i.d.\ evaluation.

Our main contributions are:

  \textbf{(1)} We design \beluga, an analysis method to measure the effect of behavioural learning. 
  It handles different kinds of behaviour measures, operationalised by labelled or perturbation-based tests.
  To that end we propose loss functions that optimise the expected behaviour of three test types: minimum functionality, invariance and directional expectation tests \cite{ribeiro_etal_checklist_ACL_2020}.

  \textbf{(2)} We extend previous work on behavioural learning by exploring two training configurations in addition to fine-tuning on suite data \cite{luzdearaujo2022checking,liu_etal_InocFineTune_NACACL_2019}: training on a mixture of i.i.d.\ and suite data; and training on i.i.d.\ data followed by fine-tuning on the data mixture.

  \textbf{(3)} We design aggregate metrics that measure generalisation across axes of different levels of granularity. From finer to coarser: generalisation within functionalities, to different functionalities and to different functionality classes.

  \textbf{(4)} We compare the generalisation capabilities of a range of regularisation techniques and domain generalisation algorithms for three representative NLP tasks (sentiment analysis, paraphrase identification and reading comprehension).

This work is not a recommendation to train on behavioural test data, but an exploration of what happens if data targeting the same set of phenomena as the tests is used for model training.
We find that naive optimisation and evaluation do yield over-optimistic scenarios: fine-tuning on suite data results in large improvements for seen functionalities, though at the same time i.i.d.\ data and unseen functionalities performance can degrade, with some models adopting degenerate solutions that pass the tests but lead to catastrophic i.i.d.\ performance.
  Including i.i.d.\ as well as test suite samples was found to prevent this, mitigating i.i.d.\ performance degradation--- with even improvements in particular cases---and yielding higher scores for unseen functionalities as well.

\section{Background}
\subsection{Behavioural testing}
\label{sec:behTesting}

% Formalize approach 
% Math defs for groups notation.

We consider a joint distribution $\p$ over an input space $\X$, corresponding label space $\Y$ and assume access to an i.i.d.\ dataset $\D$ composed of $n$ examples $\D = \{(\x_i, \y_i) \sim\p\}_{i=1}^n,~\x_i \in \X, \y_i \in \Y$, split into disjoint train, validation and test sets $\Dtrain$, $\Dval$ and $\Dtest$. We also assume access to a behavioural test suite \T, composed of $m$ test cases $\{l_i\}_{i=1}^m$ partitioned into $\nfunc$ disjoint functionalities $\{\F_i\}_{i=1}^{\nfunc}$. Each functionality belongs to one of \nclass functionality classes $\{\C_i\}_{i=1}^{\nclass}$, such that $\nclass < \nfunc < m$.

% different test types
% Formalise them
Each test case belongs to a functionality, $t \in \F_i$, and is described by a pair $(X, b)$, where $X$ is a list with $\left|X\right|$ inputs.
The expectation function $b: \mathbb{R}^{{\left|X\right|}\times\left|\Y\right|} \rightarrow \{0, 1\}$ takes a model's predictions for all  $\left|X\right|$ inputs and outputs $1$ if the model behaves as expected and $0$ otherwise.

The above taxonomy, by \citet{rottger_etal_hateCheck_ACL_2021}, describes the hierarchy of concepts in behavioural testing: functionality classes correspond to coarse properties (e.g., negation) and are composed of finer-grained functionalities; these assess facets of the coarse property (e.g., negation of positive sentiment should be negative) and are operationalised by individual input-output pairs, the test cases.
These concepts align with two of the generalisation axes we explore in this work, functionality and functionality class generalisation (§~\ref{sec:crossFunc}).

We additionally follow the terminology created by \citet{ribeiro_etal_checklist_ACL_2020}, which defines three test types, according to their evaluation mechanism: Minimum Functionality, Invariance and Directional Expectation tests.
When used for model training, each of them requires a particular optimisation strategy (§~\ref{sec:behavOptim}).

\textbf{Minimum Functionality test (MFT)}:~MFTs are input-label pairs designed to check specific system behaviour: $X$ has only one element, \x, and the expectation function checks if the model output given \x is equal to some label $\y$. 
Thus, they have the same form as the i.i.d.\ examples.

\textbf{Invariance test (INV)}:~INVs are designed to check for invariance to certain input transformations. 
The input list $X$ consists of an original input $\x_o$ and $\left|X\right|-1$ perturbed inputs $(\x_i)_{i=1}^{\left|X\right|-1}$ obtained by applying label-preserving transformations on $\x_o$.
Given model predictions $\hat{Y} := [\mathbf{\hat{y}}_i]_{i=0}^{\left|X\right|-1}$ for all inputs in $X$, then $b(\hat{Y}) = 1$ if:
\begin{equation}\label{eq:inv}
  \argmax \mathbf{\hat{y}}_0 = \argmax \mathbf{\hat{y}}_i\,,
\end{equation} 
for all $i \in \{1,\dots,\left|X\right|-1\}$. 
That is, the expectation function checks if model predictions are invariant to the perturbations. 

% \begin{equation}
%   y_o := \argmax\limits_{\y} P\left(\y|\x_o, \hat{\theta}\right)\,.
% \end{equation}

\textbf{Directional Expectation test (DIR)}:~The form for input $X$ is similar to the INV case, but instead of label-preserving transformations, $\x_o$ is perturbed in a way that changes the prediction in a task-dependent predictable way, e.g.\ prediction confidence should not increase.
Given a task-dependent comparison function $\delta: \mathbb{R}^{\left|\Y\right|}\times \mathbb{R}^{\left|\Y\right|} \rightarrow \{0,1\}$, $b(\hat{Y}) = 1$ if:
\begin{equation}
  \delta\left(\mathbf{\hat{y}}_0, \mathbf{\hat{y}}_1\right) \land \delta\left(\mathbf{\hat{y}}_0, \mathbf{\hat{y}}_2\right) \land \dots \land \delta\left(\mathbf{\hat{y}}_0, \mathbf{\hat{y}}_{\left|x\right|-1}\right)\,.
\end{equation} 
For example, if the expectation is that prediction confidence should not increase, then $\delta(\mathbf{\hat{y}}_0, \mathbf{\hat{y}}_i) = 1$ if $ \mathbf{\hat{y}}_i[c*] \leq  \mathbf{\hat{y}}_0[c*]$, where $c* := \argmax  \mathbf{\hat{y}}_0$ and $\mathbf{\hat{y}}[c*]$ denotes
the predicted probability for class $c*$.

\textbf{Evaluation}:~Given a model family $\Theta$ and a loss function $\loss: \Theta \times (\X \times \Y) \rightarrow \R_+$, the standard learning goal is to find the model $\hat{\theta} \in \Theta$ that minimises the loss over the training examples:
\begin{equation}\label{eq:erm}
  \hat{\theta} := \argmin\limits_{\theta \in \Theta}\frac{1}{\left|\Dtrain\right|} \sum\limits_{(\x, \y) \in \Dtrain} \loss (\theta, (\x, \y))\,.
\end{equation}
Then, general model correctness is evaluated using one or more metrics over the examples in $\Dtest$.
The model can be additionally evaluated using test suite \T, which gives a finer-grained performance measure over each functionality.

\subsection{Behavioural learning}
\label{sec:behLearn}

%We following the paradigm of behavioural learning \cite{luzdearaujo2022checking} and experiment with using samples from \T for training. 
In behavioural learning, samples from \T are used for training in a two-step approach: a pre-trained language model (PLM) \cite{devlin_etal_bert_NACACL_2019} is first fine-tuned on examples from \Dtrain, and then fine-tuned further on examples from \T \cite{luzdearaujo2022checking,liu_etal_InocFineTune_NACACL_2019}.

\section{\beluga}
\beluga is an analysis method to estimate how training on test suite data impacts generalisation to seen and unseen phenomena.
Given an i.i.d.\ dataset \D, a test suite \T, and a training configuration $\chi$ (§~\ref{sec:trainConf}), \beluga trains on several controlled splits of suite data and outputs scores that use performance on unseen phenomena as a proxy measure (§~\ref{sec:crossFunc}) for generalisation.

That is, \beluga can be formalised as a function $f$ parametrised by \D, \T, and $\chi$ that returns a set of metrics $M$:

\begin{equation}
  M = f(\D, \T, \chi)\,.
\end{equation}
By including measures of performance on i.i.d.\ data and on seen and unseen sets of phenomena, these metrics offer a more comprehensive and realistic view of how the training data affected model capabilities and shed light on failure cases that would be obfuscated by other evaluation schemes.

% Below we elaborate on the concepts beluga needs and how beluga works
Below we describe the examined training configurations (§~\ref{sec:trainConf}), how \beluga optimises the expected behaviours encoded in \T (§~\ref{sec:behavOptim}), how it estimates generalisation (§~\ref{sec:crossFunc}), and the metrics it outputs (§~\ref{sec:metrics}).

\subsection{Training configurations}
\label{sec:trainConf}
We split \T into three disjoint splits \Ttrain, \Tval and \Ttest, such that each split contains cases from all functionalities, and define four training configurations regarding whether and how we use \Ttrain:
% \begin{align*}
%   \Ttrain &= \bigcup\limits_{i=1}^{\nfunc}\F_{\text{train}_i}\\
%    \Tval &= \bigcup\limits_{i=1}^{\nfunc}\F_{\text{val}_i}\\
%    \Ttest &= \bigcup\limits_{i=1}^{\nfunc}\F_{\text{test}_i}\\
%    \F_i &= \F_{\text{train}_i} \cup \F_{\text{val}_i} \cup \F_{\text{test}_i},~\forall \F_i \in \T\\
%    \emptyset &= \F_{\text{train}_i} \cap \F_{\text{val}_i} \cap \F_{\text{test}_i},~\forall \F_i \in \T
% \end{align*}

  \textbf{IID}:~The standard training approach that uses only i.i.d.\ data for training (\Dtrain). It serves as a baseline to contrast performance of the three following \emph{suite-augmented} configurations.

  \textbf{\iidAndT}:~A two-step approach where first the PLM is fine-tuned on \Dtrain and then on \Ttrain. This is the setting examined in prior work on behavioural learning (§ \ref{sec:behLearn}), which has been shown to lead to deterioration of i.i.d.\ dataset (\Dtest) performance \cite{luzdearaujo2022checking}.
  
  To assess the impact of including i.i.d.\ samples in the behavioural learning procedure, we define two additional configurations:

  \textbf{\iidPlusT}:~The PLM is fine-tuned on a mixture of suite and i.i.d.\ data ($\Dtrain \cup \Ttrain$).

  \textbf{\iidAndIidPlusT}:~The PLM is first fine-tuned on \Dtrain and then on $\Dtrain \cup \Ttrain$.

By contrasting the performance on \Dtest and \Ttest of these configurations, we assess the impact of behavioural learning on both i.i.d.\ and test suite data distributions.

\subsection{Behaviour optimisation}
\label{sec:behavOptim}
Since each test type describes and expects different behaviour, \beluga optimises type-specific loss functions:

\textbf{MFT}:~As MFTs are formally equivalent to i.i.d.\ data (input-label pairs), they are treated as such: we randomly divide them into mini-batches and optimise the cross-entropy between model predictions and labels.

\textbf{INV}:~We randomly divide INVs into mini-batches composed of unperturbed-perturbed input pairs.
For each training update, we randomly select one perturbed version (of several possible) for each original input.\footnote{Note that any amount of perturbed inputs could be used, but using only one allows fitting more test cases in a mini-batch if its size is kept constant.}
We enforce invariance by minimising the cross-entropy between model predictions over perturbed-unperturbed input pairs:
\begin{equation}
  \loss(\mathbf{\hat{y}}_0, \mathbf{\hat{y}}_i) := -\sum\limits_{k=1}^c \mathbf{\hat{y}}_0[k]\cdot\log\left(\mathbf{\hat{y}}_i[k]\right)\,,
\end{equation}
where $c$ is the number of classes.
This penalises models that are not invariant to the perturbations (Eq.~\ref{eq:inv}), since the global minimum of the loss is the point where the predictions are the same.

\textbf{DIR}:~Batch construction follows the INV procedure: the DIRs are randomly divided into mini-batches of unperturbed-perturbed input pairs, the unperturbed input is randomly sampled during training.

The optimisation objective depends on the comparison function $\delta$. For a given $\delta$, we define a corresponding error measure ${\epsilon_\delta:  \mathbb{R}^{\left|\Y\right|}\times \mathbb{R}^{\left|\Y\right|}  \rightarrow [0, 1]}$. For example, if the expectation is that prediction confidence should not increase, then ${\epsilon_{\delta}(\mathbf{\hat{y}}_0, \mathbf{\hat{y}}_i) = \max\left(0, \mathbf{\hat{y}}_i[c*] - \mathbf{\hat{y}}_0[c*]\right)}$. This way, $\epsilon_\delta$ increases with confidence increase and is zero otherwise.

We minimise the following loss:
\begin{equation}
  \loss(\mathbf{\hat{y}}_0, \mathbf{\hat{y}}_i, \delta) := -\log \left(1 - \epsilon_\delta(\mathbf{\hat{y}}_0, \mathbf{\hat{y}}_i)\right)\,.
\end{equation}    
Intuitively, if $\epsilon_\delta=0$, the loss is zero.
Conversely, the loss increases with the error measure (as $\epsilon_\delta$ gets closer to 1).

\subsection{Cross-functional analysis}
\label{sec:crossFunc}
% why cross functional
Test suites have limited coverage: the set of covered functionalities is only a subset of the phenomena of interest:  $\T \subset \phenom$, where $\phenom$ is the hypothetical set of all functionalities.
For example, the test suite for sentiment analysis provided by \citet{ribeiro_etal_checklist_ACL_2020} has a functionality that tests for invariance to people's names---the sentiment of the sentence ``I do not like Mary's favourite movie'' should not change if ``Mary'' is changed to ``Maria''.
However, the equally valid functionality that tests for invariance to organisations' names is not in the suite.
Training and evaluating on the same set of functionalities can lead to overestimating the performance: models that overfit to covered functionalities but fail catastrophically on non-covered ones.

\beluga computes several measures of model performance that address  generalisation from \Ttrain to \Ttest and from $\Ttrain$ to  $\phenom$.
We do not assume access to test cases for non-covered phenomena, so we use held-out sets of functionalities as proxies for generalisation to \phenom.
% \added{
% To that end, we employ a cross-functional analysis framework, brought up by \mbox{\citet{luzdearaujo2022checking}}, based on training and evaluating on different sets of phenomena (e.g., functionalities or functionality classes).
% Our approach differs from theirs in two ways.
% First, they use overall accuracy as the evaluation metric, which implicitly assign less importance to functionalities with less samples, and thus obfuscates performance change on underrepresented functionalities. 
% Our metrics assign equal importance to all functionalities.
% Second, they use different train-test splits for each training run, which introduces sample variability as a confounder: a drop in performance can be due to a less accurate model, a more difficult test set,  or some combination of both factors.
% We use a fixed train-test split, varying only which sets of functionalities the model has access during training.
% We describe below how \beluga evaluates i.i.d.\ and test suite data:}

\textbf{I.i.d.\ data}:~To score performance on \Dtest, we use the canonical evaluation metric for the specific \emph{dataset}. We detail the metrics used for each examined \emph{task}\footnote{We refer to the i.i.d.\ data as the \emph{dataset} as opposed to the \emph{task}. The task is more abstract, and it comes with a corresponding behavioural test suite.} in Section~\ref{sec:tasks}. We denote the i.i.d.\ score as $s_{iid}$.

\textbf{Test suite data}:~We compute the pass rate $s_{\F_i}$ of each functionality $\F_i\in \T$:
\begin{equation}\label{eq:pass}
  s_{\F_i} := \frac{1}{\left|\F_{\text{test}_i}\right|}\sum\limits_{(X, b) \in \F_{\text{test}_i}} b(\hat{Y})\,,
\end{equation}
where $\hat{Y}$ are the model prediction given the inputs in $X$. In other words, the pass rate is simply the proportion of successful test cases.

We vary the set of functionalities used for training and testing to construct different evaluation scenarios:

\textbf{Unseen evaluation}:~No test cases are seen during training.
This is equivalent to the use of behavioural test suites without behavioural learning: we compute the pass rates using the predictions of an IID model.

\textbf{Seen evaluation}:~\Ttrain is used for training.
We compute the pass rate on \Ttest using the predictions of suite-augmented models.
This score measures how well the fine-tuning procedure generalises to test cases of \textit{covered} functionalities: even though all functionalities are seen during training, the particular test cases evaluated (${\{t|t\in \Ttest\}}$) are not the same as the ones used for training ($\Ttrain \cap \Ttest = \emptyset$).

\textbf{Generalisation to non-covered phenomena}:~To estimate performance on non-covered phenomena, we construct a $l$-subset partition of the set of functionalities $U := \{U_i\}_{i=1}^{l}$. For each $U_i$, we use $\Ttrain \setminus U_i$ for training and then compute the pass rates for $\Ttest \cap U_i$: $\{s_{\F\text{unseen}} | \F \in U_i\}$. That is, we fine-tune it on a set of functionalities and evaluate it on the remaining (unseen) functionalities. Since $U$ is a partition of \T, by the end of the procedure there will be a pass rate for each functionality.

We consider three different partitions, depending on the considered generalisation proxy:

% splits
% Math defs for the splits, seen vs unseen sets.
  \textbf{(1)} Functionality generalisation: a partition with $\nfunc$ subsets, each corresponding to a held-out functionality: $U_i = \{\F_i\},~i \in \{1,\dots,\nfunc\}$. 
  % That is, for each functionality, we compute its pass rate using a model fine-tuned on all the other functionalities. 
  We consider this a proxy of performance on non-covered functionalities: $\F \in \phenom \setminus \T$.

  \textbf{(2)} Functionality class generalisation: a partition with $\nclass$ subsets, each corresponding to a held-out functionality class: $U_i = \{\C_i\},~i \in \{1,\dots,\nclass\}$. 
  % That is, for each functionality class, we compute the pass rates for its functionalities using a model fine-tuned on all the other classes. 
  We consider this to be a proxy of performance on non-covered functionality classes: $\C \subset \phenom \setminus \T$.

  \textbf{(3)} Test type generalisation: a partition with three subsets, each corresponding to a held-out test type: $U_i = \{\F | \F~\text{has type }i\},~i \in \{\text{MFT},\text{INV},\text{DIR}\}$. 
  % That is, for each test type, we compute the pass rates for the functionalities that have are of such type using a model fine-tuned on all the other types. 
  We use this measure to examine generalisation across different test types.

% Aggregate with iid results
\subsection{Metrics}
\label{sec:metrics}
For model comparison purposes, \beluga outputs the average pass rate (the arithmetic mean of the $\nfunc$ pass rates) as the aggregated metric for test suite correctness.
Since one of the motivations for behavioural testing is its fine-grained results, \beluga also reports the individual pass rates.

% One possible method would be to consider the worst-case accuracy by using the lowest pass rate as the aggregate metric. 
% %This makes sense when we consider that models can have 
% This could be intuitive for sensitive applications where the worst-case performance can produce disastrous consequences.
% However, such metric would not be able to discriminate models. 
% For example, in our experiments, it was common for models to fail all test cases for a given functionality. 
% In such cases, two models $\theta_a$ and $\theta_b$ would score 0, even if one performed much better than the other on most functionalities.

In total, \beluga computes five aggregated suite scores, each corresponding to an evaluation scenario: 

   $s_{\T \text{standard}}$: The baseline score of a model only trained on i.i.d.\ data: if the other scores are lower, then fine-tuning on test suite data degraded overall model performance.

   $s_{\T \text{seen}}$: Performance on seen functionalities. This score can give a false sense of model performance since it does not account for model overfitting to the seen functionalities: spurious correlations within functionalities and functionality classes can be exploited to get deceivingly high scores.

   $s_{\T \text{func}}$: Measure of generalisation to unseen functionalities. It is a more realistic measure of model quality, but since functionalities correlate within a functionality class, the score may still offer a false sense of quality.

    $s_{\T \text{class}}$: Measure of generalisation to unseen functionality classes. This is the most challenging generalisation setting, as the model cannot exploit correlations within functionalities and functionality classes.

   $s_{\T \text{type}}$: Measure of generalisation to unseen test types. This score is of a more technical interest: it can offer insights into how different training signals affect each other (e.g.\ if training with MFTs supports performance on INVs and vice-versa).

\textbf{Comprehensive generalisation score}:~
Since performance on i.i.d.\ data and passing the behavioural tests are both important, \beluga provides the harmonic mean of the aggregated pass rates and the i.i.d.\ score as an additional metric for model comparison:
\begin{equation}\label{eq:g}
  \g :=  2\frac{s_{\T}\cdot s_{iid}}{s_{\T} + s_{iid}}\,.
\end{equation}
%One alternative would be to just take the geometric mean of all pass rates and the i.i.d.\ score, but this would assign to the i.i.d.\ score the same importance as individual functionality scores. 
%Conversely, we consider i.i.d.\ performance to be as important as performance on the test suite as a whole, so we opt for the formulation given by Eq.~\ref{eq:g}.
There are five \g scores (\gstandard, \gseen, \gfunc, \gclass and \gtype), each corresponding to plugging either $s_{\T \text{standard}}$, $s_{\T \text{seen}}$, $s_{\T \text{func}}$, $s_{\T \text{class}}$ or $s_{\T \text{type}}$ into Eq.~\ref{eq:g}.

This aggregation makes implicit importance assignments explicit: on the one hand, the harmonic mean  ensures that both i.i.d.\ and suite performance are important due to its sensitivity to low scores; on the other, different phenomena are weighted differently, as i.i.d.\ performance has a bigger influence on the final score than each single functionality pass rate.

\section{Experiments on cross-functional analysis}
% Experimental setting.
\begin{table*}[tb]
  \centering
  \footnotesize
  \rowcolors{2}{gray!20}{white}
  \begin{tabularx}{\linewidth}{lX}\toprule
    \rowcolor{white}
    \makecell[l]{Dataset} &Example (label)\\\midrule
SST-2 &A sensitive, moving ,brilliantly constructed work. (Positive) \\
&By far the worst movie of the year. (Negative) \\
QQP  &Q1: Who is king of sports? Q2:Who is the king? (Not duplicate) \\
&Q1: How much does it cost to build an basic Android app in India? Q2: How much does it cost to build an Android app in India? (Duplicate) \\
SQuAD&C: Solar energy may be used in a water stabilisation pond to treat waste [...] although algae may produce toxic chemicals that make the water unusable. Q: What is a reason why the water from a water stabilisation pond may be unusable? (algae may produce  toxic chemicals)\\
    \bottomrule
  \end{tabularx}
  \caption{Examples for each i.i.d.\ dataset. The number of train/validation/test samples is 67k/436/436, 363k/20k/20k and 87k/5k/5k for SST-2, QQP and SQuAD, respectively.}\label{tab:iidData}
  \end{table*}

\begin{table*}[tb]
  \centering
  \footnotesize
  \rowcolors{2}{gray!25}{white}
  \begin{tabularx}{\textwidth}{lXX}\toprule
    \rowcolor{white}
    Task &Example input (expected behaviour)  &Class---Functionality (type)\\\midrule
    % \texttt{SENT}  &The service is not creepy. &Negative &Negation & Simple negations: not negative & MFT  \\
    \texttt{SENT}&  \textcolor{PineGreen}{I used to think} this is an incredible food. (Not more confident) &Temporal---Prepending ``I used to think'' to a statement should not raise prediction confidence
    (DIR)  \\
    & Hannah is a \textcolor{red}{Christian} $\rightarrow$ \textcolor{PineGreen}{Buddhist} model. (Same prediction)&Fairness---Prediction should be invariant to religion identifiers (INV)  \\
    % & &@SouthwestAir Thank you for the tip! \textcolor{PineGreen}{Never flying with you again.}&Not less negative &Vocabulary & Add negative phrases & DIR \\
    \texttt{PARA} &  Q1: Are tigers heavier than computers? Q2: What is heavier, computers or tigers? (Duplicate) & SRL---Changing comparison order preserves question semantics (MFT)  \\
    % &  Q1: Can \textcolor{red}{India} $\rightarrow$ \textcolor{PineGreen}{South Africa}  and Pakistan be friends? Q2: Why should \textcolor{red}{India} $\rightarrow$ \textcolor{PineGreen}{South Africa} and Pakistan be friends? &Same prediction &NER & Change same location in both questions & INV \\
    & Q1: What are the best venture capital firms in \textcolor{red}{India} $\rightarrow$ \textcolor{PineGreen}{Albania}? Q2: Which is the first venture capital firm in India? (Not duplicate) &NER---Questions referring to different locations are not duplicate (DIR)  \\
    \texttt{READ}&C: Somewhere around a billion years ago, a free-living cyanobacterium entered an early eukaryotic cell [...]  Q: \textcolor{red}{What kind} $\rightarrow$ \textcolor{PineGreen}{Wha tkind} of cell did cyanobacteria enter long ago? (Same prediction) &Robustness---Typos should not change prediction (INV) \\
     & C: Maria is an intern. Austin is an editor. Q: Who is not an intern? (Austin) &Negation---Negations in question matter for prediction (MFT)\\
    \bottomrule
  \end{tabularx}
  \caption{Examples for each test suite. We color-code perturbations as red/green for deletions/additions. The number of train/validation/test samples is 89k/44k/44k, 103k/51k/51k and 35k/17k/17k for the \texttt{SENT}, \texttt{PARA} and \texttt{READ} test suites, respectively.}\label{tab:tData}
  \end{table*}

% tasks
\subsection{Tasks}
\label{sec:tasks}
We experiment with three classification tasks that correspond to the test suites made available\footnote{\url{https://github.com/marcotcr/checklist}.} by \citet{ribeiro_etal_checklist_ACL_2020}: sentiment analysis (\texttt{SENT}), paraphrase identification (\texttt{PARA}) and reading comprehension (\texttt{READ}).\footnote{These test suites were originally proposed for model evaluation.
Every design choice we describe regarding optimisation (e.g.\ loss functions and label encodings) is ours.}
Tables \ref{tab:iidData} and \ref{tab:tData} summarise and show representative examples from the i.i.d.\ and test suite datasets, respectively.

  % Fixing qqp wrong label
  % ensure that samples from the same test case are in the same split (check notebooks when describing approach)

\textbf{Sentiment analysis (\texttt{SENT})}:~As the i.i.d.\ dataset for sentiment analysis, we use the Stanford Sentiment Treebank (SST-2) \cite{socher2013recursive}. 
We use the version made available in the GLUE benchmark \cite{wang2018glue}, where the task is to assign binary labels (negative/positive sentiment) to sentences. 
The test set labels are not publicly available, so we split the original validation set in half as our validation and test sets. 
The canonical metric for the dataset is accuracy.

The \texttt{SENT} suite contains 68k MFTs, 9k DIRs and 8k INVs. 
It covers functionality classes such as semantic role labelling (SRL), named entity recognition (NER) and fairness. 
The MFTs were template-generated, while the DIRs and INVs were either template-generated or obtained from perturbing a dataset of unlabelled airline tweets. 
Therefore, there is a domain mismatch between the i.i.d.\ data (movie reviews) and the suite data (tweets about airlines).

There are also label mismatches between the two datasets: the suite contains an additional class for neutral sentiment and the MFTs have the ``not negative'' label, which admits both positive and neutral predictions.
We follow \citet{ribeiro_etal_checklist_ACL_2020} and consider predictions with probability of positive sentiment within $[1/3, 2/3]$ as neutral.\footnote{When training, we encode ``neutral'' and ``not negative'' labels as  $[1/2, 1/2]$ and $[1/3, 2/3]$, respectively.
One alternative is to create two additional classes for such cases, but this would prevent the use of the classification head fine-tuned on i.i.d.\ data (which is annotated with binary labels).}

There are two types of comparison for DIRs, regarding either sentiment or prediction confidence.
In the former case, the prediction for a perturbed input is expected to be either not more negative or not more positive when compared with the prediction for the original input.
In the latter, the confidence of the original prediction is expected to either not increase or not decrease, regardless of the sentiment.
For example, when adding an intensifier (``really'', ``very'') or a reducer (``a little'', ``somewhat''), the confidence of the original prediction should not decrease in the first case and not increase in the second. 
On the other hand, if a perturbation adds a positive or negative phrase to the original input, the positive probability should not go down (up) for the first (second) case. 

More formally, each prediction $\mathbf{\hat{y}}$ is a two-dimensional vector where the first and second components are the confidence for negative ($\mathbf{\hat{y}}[0]$) and positive ($\mathbf{\hat{y}}[1]$) sentiment, respectively. Let $c*$ denote the component with highest confidence in the \textit{original} prediction: $c* := \argmax\mathbf{\hat{y}}_0$. Then, the comparison function $\delta$ can take one of four forms (not more negative, not more positive, not more confident and not less confident):
\begin{align*}
   \delta_{\uparrow p}(\mathbf{\hat{y}}_0, \mathbf{\hat{y}}_i)&=1 ~\text{if}~ \mathbf{\hat{y}}_i[0] \leq \mathbf{\hat{y}}_0[0]\\
   \delta_{\uparrow n}(\mathbf{\hat{y}}_0, \mathbf{\hat{y}}_i)&=1 ~\text{if}~ \mathbf{\hat{y}}_i[1] \leq \mathbf{\hat{y}}_0[1]\\
   \delta_{\downarrow c}(\mathbf{\hat{y}}_0, \mathbf{\hat{y}}_i)&=1 ~\text{if}~ \mathbf{\hat{y}}_i[c*] \leq \mathbf{\hat{y}}_0[c*]\\
   \delta_{\uparrow c}(\mathbf{\hat{y}}_0, \mathbf{\hat{y}}_i)&=1 ~\text{if}~ \mathbf{\hat{y}}_i[c*] \geq \mathbf{\hat{y}}_0[c*]
\end{align*}
Each corresponding to an error measure $\epsilon$:
\begin{align*}
  \epsilon_{\delta_{\uparrow p}}(\mathbf{\hat{y}}_0, \mathbf{\hat{y}}_i) &:= \max\left(0, \mathbf{\hat{y}}_i[0] - \mathbf{\hat{y}}_0[0]\right) \\
  \epsilon_{\delta_{\uparrow n}}(\mathbf{\hat{y}}_0, \mathbf{\hat{y}}_i) &:= \max\left(0, \mathbf{\hat{y}}_i[1] - \mathbf{\hat{y}}_0[1]\right) \\
  \epsilon_{\delta_{\downarrow c}}(\mathbf{\hat{y}}_0, \mathbf{\hat{y}}_i) &:= \max\left(0, \mathbf{\hat{y}}_i[c*] - \mathbf{\hat{y}}_0[c*]\right) \\
  \epsilon_{\delta_{\uparrow c}}(\mathbf{\hat{y}}_0, \mathbf{\hat{y}}_i) &:= \max\left(0, \mathbf{\hat{y}}_0[c*] - \mathbf{\hat{y}}_i[c*]\right)
\end{align*}
We compute the $\max$ because only test violations should be penalised.
% For example, in the ``not more negative'' case, it is ok if the negative class confidence is reduced after the perturbation.
% In this case the difference will be negative, so the $\max$ function will yield zero error.
% The reasoning is analogue for the other three cases.

\textbf{Paraphrase identification (\texttt{PARA})}:~We use Quora Question Pairs (QQP) \cite{iyer2017quora} as the i.i.d.\ dataset.
It is composed of question pairs from the website Quora with annotation for whether a pair of questions is semantically equivalent (duplicates or not duplicates). 
% We use the dataset version packaged in the GLUE benchmark, where the task is to classify the pairs as duplicates (semantically equivalent) or not. 
The test set labels are not available, hence we split the original validation set into two sets for validation and testing. 
The canonical metrics are accuracy and the \fscore score of the duplicate class.

The \texttt{PARA} suite contains 46k MFTs, 13k DIRs and 3k INVs, with functionality classes such as co-reference resolution, logic and negation. 
All MFTs are template generated,\footnote{The test cases from functionality ``Order does matter for asymmetric relations'' (e.g.\ Q1: Is Rachel faithful to Christian?, Q2: Is Christian faithful to Rachel?) were originally labelled as duplicates.
This seems to be unintended, so we change their label to not duplicates.} while the INVs and DIRs are obtained from perturbing QQP data.

The DIRs are similar to MFTs: perturbed question pairs are either duplicate or not duplicate. 
For example, if two questions mention the same location and the perturbation changes the location in one of them, then the new pair is guaranteed not to be semantically equivalent. 
Thus, the comparison function $\delta$ checks if the perturbed predictions correspond to the expected label; the original prediction is not used for evaluation.
So during training, we treat them as MFTs: we construct mini-batches of perturbed samples and corresponding labels and minimise the cross-entropy between predictions and labels.

% Let the duplicate and not duplicate labels be denoted by the integers 1 and 0. 
% Then $\delta$ takes on of two forms (duplicate and not duplicate expectation):
% \begin{enumerate}
%   \item $\delta_1(\hat{y}_0, \hat{y}_i)=1$ if $\hat{y}_i = 1$: duplicate expectation.
%   \item $\delta_0(\hat{y}_0, \hat{y}_i)=1$ if $\hat{y}_i = 0$: not duplicate expectation.
% \end{enumerate}
% Each of them corresponding to an error measure $\epsilon$:
% \begin{align*}
%   \epsilon_{\delta_{\uparrow p}}(\hat{y}_0, \hat{y}_i) &:= \max\left(0, \hat{y}_i[0] - \hat{y}_0[0]\right) \\
%   \epsilon_{\delta_{\uparrow n}}(\hat{y}_0, \hat{y}_i) &:= \max\left(0, \hat{y}_i[1] - \hat{y}_0[1]\right) \\
%   \epsilon_{\delta_{\downarrow c}}(\hat{y}_0, \hat{y}_i) &:= \max\left(0, \hat{y}_i[c*] - \hat{y}_0[c*]\right) \\
%   \epsilon_{\delta_{\uparrow c}}(\hat{y}_0, \hat{y}_i) &:= \max\left(0, \hat{y}_0[c*] - \hat{y}_i[c*]\right)
% \end{align*}
% Therefore, the original prediction does not matter for these tests.

\textbf{Reading comprehension (\texttt{READ})}:~The i.i.d.\ dataset for \texttt{READ} is the Stanford Question Answering Dataset (SQuAD) \cite{rajpurkar2016squad}, composed of excerpts from Wikipedia articles with crowdsourced questions and answers.
The task is to, given a text passage (context) and a question about it, extract the context span that contains the answer.
Once again, the test set labels are not publicly available and we repeat our splitting approach for \texttt{SENT} and \texttt{PARA}.
The canonical metrics are exact string match (EM) (percentage of predictions that match ground truth answers exactly) and the more lenient \fscore score, which measures average token overlap between predictions and ground truth answers.

The \texttt{READ} suite contains 10k MFTs and 2k INVs, with functionality classes such as vocabulary and taxonomy.
The MFTs are template generated, while the INVs are obtained from perturbing SQuAD data.

Invariance training in \texttt{READ} has one complication, since the task is to extract the answer span by predicting the start and end positions.
Naively using the originally predicted positions would not work because the answer position may have changed after the perturbation. 
For example, let us take the original context-question pair (C: Paul travelled from Chicago to New York, Q: Where did Paul travel to?) and perturb it so that Chicago is changed to Los Angeles.
The correct answer for the original input is (5, 6) as the start and end (word) positions, yielding the span ``New York''.
Applying these positions to the perturbed input would extract ``to New''.
% It would not make sense to compare the predictions for these two spans.
% Of course, one could identify the originally predicted span, find it in the perturbed version and then minimise the cross-entropy between the originally predicted probability for the start (and end) position and the predicted probability for the equivalent position in the perturbed input.
% But such matching is not easy to vectorize and would consequently slow down training. 
Instead, we only compare the model outputs for the positions that correspond to the common ground of original and perturbed inputs.
In the example, the outputs for the tokens ``Paul'', ``travelled'', ``from'', ``to'', ``New'' and ``York''.
We minimise the cross-entropy between this restricted set of outputs for the original and perturbed inputs.
This penalises changes in prediction for equivalent tokens (e.g.\ the probability of ``Paul'' being the start of the answer is $0.1$ for the original input but $0.15$ for the perturbed).

\subsection{Generalisation methods}
We use \beluga to compare several techniques used to improve generalisation:

\textbf{L2}:~We apply a stronger-than-typical $\ell_2$-penalty coefficient of $\lambda=0.1$.

\textbf{Dropout}:~We triple the dropout rate for all fully connected layers and attention probabilities from the default value of $0.1$ to $0.3$.

\textbf{LP}:~Instead of fine-tuning on suite data, we apply linear probing (LP), where the encoder parameters are frozen, and only the classification head parameters are updated.
Previous work \cite{kumar2022finetuning} has found this to generalise better than full fine-tuning.

\textbf{LP-FT}:~We experiment with linear probing followed by fine-tuning, which \citet{kumar2022finetuning} have shown to combine the benefits of fine-tuning (in-distribution performance) and linear-probing (out-of-distribution performance).

\textbf{Invariant risk minimisation (IRM)} \cite{arjovsky2019invariant}, a framework for OOD generalisation that leverages different training environments to learn feature-label correlations that are invariant across the environments, under the assumption that such features are not spuriously correlated with the labels.

\textbf{Group distributionally robust optimisation (Group-DRO)} \cite{Sagawa2020Distributionally}, an algorithm that minimises not the average training loss, but the highest loss across the different training environments. This is assumed to prevent the model from adopting spurious correlations as long as such correlations do not hold on one of the environments.

\textbf{Fish} \cite{shi_etal_gradientMatchDomainGen_arxiv_2021}, an algorithm for domain generalisation that maximises the inner product between gradients from different training environments, under the assumption that this leads models to learn features invariant across environments.

For the last three methods, we treat the different functionalities as different environments. For the \iidPlusT and \iidAndIidPlusT settings, we consider the i.i.d.\ data as an additional environment.
In the multi-step training configurations (\iidAndT and \iidAndIidPlusT), we only apply the techniques during the second step: when training only with i.i.d.\ data we employ vanilla gradient descent, since we are interested in the generalisation effect of using suite data.

\subsection{Experimental setting}
% Start from BERT-BASE or bert-large
% oPTMISER
% Hyperparameter tuning
% Hyperparameters
\label{sec:settings}
We use pre-trained BERT models \cite{devlin_etal_bert_NACACL_2019} for all tasks.
We follow \citet{ribeiro_etal_checklist_ACL_2020} and use BERT-base for \texttt{SENT} and \texttt{PARA} and BERT-large for \texttt{READ}.
All our experiments use AdamW \cite{loschilov2019adamw} as the optimiser. When fine-tuning on i.i.d.\ data, we use the same hyper-parameters as the ones reported for models available on Hugging Face's model zoo.\footnote{Available on \url{https://huggingface.co/}. The model names are textattack/bert-base-uncased-SST-2 (\texttt{SENT}), textattack/bert-base-uncased-QQP (\texttt{PARA}) and bert-large-uncased-whole-word-masking-finetuned-squad (\texttt{READ}).}
When fine-tuning on test suite data, we run a grid search over a range of values for batch size, learning rate and number of epochs.\footnote{Batch size:$\{2, 3\}$ for \texttt{READ} and $\{8, 16\}$ for the others; learning rate: $\{\num{2e-5}, \num{3e-5}, \num{5e-5}\}$; number of epochs: $\{1, 2, 3\}$.}
We select the configuration that performed best on \Tval.
To maintain the same compute budget across all methods, we do not tune method-specific hyper-parameters.
We instead use values shown to work well in the original papers and previous work \cite{dranker2021irmwhen}.

\section{Results and observations}
\begin{table*}[tb]
  \footnotesize
  \centering
  \begin{tabularx}{\textwidth}{llXXXXXXXXXXXXXXXX}
    \toprule
    Config & Method & SST2 & QQP & SQuAD &\multicolumn{4}{c}{\texttt{SENT}} &\multicolumn{4}{c}{\texttt{PARA}} &\multicolumn{4}{c}{\texttt{READ}} &  \\
    \cmidrule(lr){3-3}\cmidrule(lr){4-4}\cmidrule(lr){5-5}\cmidrule(lr){6-9} \cmidrule(lr){10-13}\cmidrule(lr){14-17}
    & & Acc. & Acc. & EM &\gseen & \gfunc &\gclass &\gtype & \gseen & \gfunc &\gclass &\gtype & \gseen & \gfunc &\gclass &\gtype & Avg. \\
    \midrule
    IID & Vanilla & 91.74 & 91.28 & 84.58 & 72.94 & 72.94 & 72.94 & 72.94 & 74.70 & 74.70 & 74.70 & 74.70 & 67.58 & 67.58 & 67.58 & \textbf{67.58} & 71.74 \\
    \midrule
    \multirow{8}{*}{\rotatebox[origin=c]{90}{\iidAndT}} & Vanilla & \textcolor{red}{82.34} & \textcolor{red}{89.36} & \textcolor{red}{3.82} & \textcolor{PineGreen}{90.31} & \textcolor{PineGreen}{86.58} & \textcolor{PineGreen}{80.95} & \textcolor{red}{65.98} & \textcolor{PineGreen}{93.29} & \textcolor{PineGreen}{80.05} & \textcolor{PineGreen}{75.75} & \textcolor{red}{73.72} & \textcolor{red}{7.33} & \textcolor{red}{7.04} & \textcolor{red}{6.86} & \textcolor{red}{6.60} & \textcolor{red}{56.21} \\
 & L2 & \textcolor{red}{78.90} & \textcolor{red}{87.70} & \textcolor{red}{0.83} & \textcolor{PineGreen}{88.17} & \textcolor{PineGreen}{84.62} & \textcolor{PineGreen}{80.51} & \textcolor{red}{68.24} & \textcolor{PineGreen}{92.34} & \textcolor{PineGreen}{75.55} & \textcolor{red}{70.93} & \textcolor{red}{71.35} & \textcolor{red}{1.65} & \textcolor{red}{1.63} & \textcolor{red}{1.63} & \textcolor{red}{1.62} & \textcolor{red}{53.19} \\
 & Dropout & \textcolor{red}{83.26} & \textcolor{red}{86.70} & \textcolor{red}{1.57} & \textcolor{PineGreen}{90.86} & \textcolor{PineGreen}{88.85} & \textcolor{PineGreen}{84.44} & \textcolor{red}{68.13} & \textcolor{PineGreen}{91.44} & \textcolor{PineGreen}{78.45} & \textcolor{red}{72.57} & \textcolor{red}{69.17} & \textcolor{red}{3.09} & \textcolor{red}{3.03} & \textcolor{red}{3.01} & \textcolor{red}{3.01} & \textcolor{red}{54.67} \\
 & LP & \textcolor{red}{86.24} & \textcolor{red}{88.70} & 84.05 & \textcolor{PineGreen}{80.98} & \textcolor{PineGreen}{77.59} & \textcolor{PineGreen}{74.49} & \textcolor{red}{65.61} & \textcolor{PineGreen}{78.84} & \textcolor{red}{74.03} & \textcolor{red}{71.50} & \textcolor{red}{69.67} & \textcolor{PineGreen}{76.11} & \textcolor{PineGreen}{68.96} & \textbf{68.09} & \textcolor{red}{65.50} & \textcolor{PineGreen}{72.61} \\
 & LP-FT & \textcolor{red}{80.28} & \textcolor{red}{90.01} & \textcolor{red}{1.15} & \textcolor{PineGreen}{89.06} & \textcolor{PineGreen}{87.11} & \textcolor{PineGreen}{84.53} & \textcolor{red}{64.40} & \textcolor{PineGreen}{93.48} & \textcolor{PineGreen}{79.87} & 75.19 & \textcolor{red}{72.58} & \textcolor{red}{2.27} & \textcolor{red}{2.25} & \textcolor{red}{2.24} & \textcolor{red}{2.23} & \textcolor{red}{54.60} \\
 & IRM & \textcolor{red}{79.36} & \textcolor{red}{88.77} & \textcolor{red}{83.05} & \textcolor{PineGreen}{88.48} & \textcolor{PineGreen}{84.42} & 73.63 & \textcolor{red}{69.18} & \textcolor{PineGreen}{92.87} & \textcolor{PineGreen}{80.51} & 74.61 & \textcolor{red}{71.58} & \textcolor{PineGreen}{90.11} & \textcolor{PineGreen}{71.36} & \textcolor{red}{66.23} & \textcolor{red}{35.90} & \textcolor{PineGreen}{74.91} \\
 & DRO & \textcolor{red}{83.72} & \textcolor{red}{82.71} & \textcolor{red}{0.61} & \textcolor{PineGreen}{91.14} & \textcolor{PineGreen}{86.56} & \textcolor{PineGreen}{78.85} & \textcolor{red}{66.11} & \textcolor{PineGreen}{89.60} & \textcolor{red}{73.58} & \textcolor{red}{69.29} & \textcolor{red}{71.73} & \textcolor{red}{1.21} & \textcolor{red}{1.20} & \textcolor{red}{1.20} & \textcolor{red}{1.20} & \textcolor{red}{52.64} \\
 & Fish & \textcolor{red}{84.63} & \textcolor{red}{88.61} & 84.03 & \textcolor{PineGreen}{91.68} & \textcolor{PineGreen}{87.22} & \textcolor{PineGreen}{74.75} & \textcolor{red}{70.84} & \textcolor{PineGreen}{92.89} & \textbf{\textcolor{PineGreen}{81.61}} & \textcolor{PineGreen}{75.91} & 74.42 & \textcolor{PineGreen}{90.61} & \textcolor{PineGreen}{68.80} & \textcolor{red}{66.00} & \textcolor{red}{65.90} & \textcolor{PineGreen}{78.39} \\
 \midrule
 \multirow{8}{*}{\rotatebox[origin=c]{90}{\iidPlusT}} & Vanilla & 91.28 & \textcolor{PineGreen}{91.87} & 85.45 & \textcolor{PineGreen}{94.15} & \textcolor{PineGreen}{90.97} & \textcolor{PineGreen}{80.07} & \textcolor{red}{71.81} & \textcolor{PineGreen}{93.98} & \textcolor{PineGreen}{77.93} & \textcolor{red}{72.63} & 75.23 & \textcolor{PineGreen}{91.35} & 66.54 & \textcolor{red}{64.40} & \textcolor{red}{63.12} & \textcolor{PineGreen}{78.52} \\
 & L2 & 89.45 & \textcolor{PineGreen}{91.80} & \textcolor{PineGreen}{86.02} & \textcolor{PineGreen}{93.49} & \textcolor{PineGreen}{88.37} & \textcolor{PineGreen}{77.98} & \textcolor{red}{70.15} & \textcolor{PineGreen}{94.20} & \textcolor{PineGreen}{78.34} & \textcolor{red}{73.27} & 74.81 & \textbf{\textcolor{PineGreen}{91.94}} & \textbf{\textcolor{PineGreen}{72.28}} & \textcolor{red}{61.88} & \textcolor{red}{63.58} & \textcolor{PineGreen}{78.36} \\
 & Dropout & 91.74 & \textcolor{red}{89.89} & 85.13 & \textbf{\textcolor{PineGreen}{95.69}} & \textcolor{PineGreen}{90.77} & \textcolor{PineGreen}{84.49} & \textbf{\textcolor{PineGreen}{74.03}} & \textcolor{PineGreen}{93.18} & 75.39 & 74.16 & 74.49 & \textcolor{PineGreen}{91.22} & 67.19 & \textcolor{red}{62.42} & \textcolor{red}{62.69} & \textcolor{PineGreen}{78.81} \\
 & LP & \textcolor{red}{78.44} & \textcolor{red}{66.50} & \textcolor{red}{16.58} & 70.96 & \textcolor{red}{68.58} & \textcolor{red}{66.29} & \textcolor{red}{67.55} & \textcolor{red}{59.95} & \textcolor{red}{58.77} & \textcolor{red}{59.84} & \textcolor{red}{59.90} & \textcolor{red}{16.77} & \textcolor{red}{16.29} & \textcolor{red}{16.17} & \textcolor{red}{15.66} & \textcolor{red}{48.06} \\
 & LP-FT & 91.28 & 91.16 & \textbf{\textcolor{PineGreen}{86.13}} & \textcolor{PineGreen}{94.14} & \textcolor{PineGreen}{89.37} & \textcolor{PineGreen}{75.31} & \textcolor{red}{71.91} & \textcolor{PineGreen}{93.90} & 75.36 & \textcolor{red}{73.48} & 74.87 & \textcolor{PineGreen}{91.65} & \textcolor{PineGreen}{72.17} & \textcolor{red}{64.64} & \textcolor{red}{62.72} & \textcolor{PineGreen}{78.29} \\
 & IRM & \textcolor{red}{57.11} & \textcolor{red}{50.59} & \textcolor{red}{10.94} & 72.70 & 70.90 & \textcolor{red}{69.08} & \textcolor{red}{64.26} & \textcolor{red}{66.30} & \textcolor{red}{50.12} & \textcolor{red}{52.63} & \textcolor{red}{51.56} & \textcolor{red}{19.50} & \textcolor{red}{11.40} & \textcolor{red}{10.73} & \textcolor{red}{10.62} & \textcolor{red}{45.82} \\
 & DRO & \textcolor{red}{86.24} & \textcolor{red}{84.28} & \textcolor{red}{74.51} & \textcolor{PineGreen}{92.61} & \textcolor{PineGreen}{89.44} & \textcolor{PineGreen}{78.99} & \textcolor{red}{67.52} & \textcolor{PineGreen}{90.43} & \textcolor{red}{72.25} & \textcolor{red}{73.09} & \textcolor{red}{67.89} & \textcolor{red}{63.21} & \textcolor{red}{50.68} & \textcolor{red}{52.06} & \textcolor{red}{54.35} & \textcolor{red}{71.04} \\
 & Fish & \textcolor{red}{87.39} & \textcolor{red}{77.64} & \textcolor{red}{70.50} & \textcolor{PineGreen}{93.27} & \textcolor{PineGreen}{89.37} & \textcolor{PineGreen}{78.58} & \textcolor{red}{70.48} & \textcolor{PineGreen}{86.20} & \textcolor{red}{62.57} & \textcolor{red}{65.22} & \textcolor{red}{71.15} & \textcolor{PineGreen}{82.01} & \textcolor{red}{58.38} & \textcolor{red}{47.68} & \textcolor{red}{56.22} & 71.76 \\

 \midrule
 \multirow{8}{*}{\rotatebox[origin=c]{90}{\iidAndIidPlusT}} & Vanilla & 90.83 & \textcolor{PineGreen}{91.79} & \textcolor{red}{83.41} & \textcolor{PineGreen}{93.92} & \textcolor{PineGreen}{90.04} & \textcolor{PineGreen}{80.35} & \textcolor{red}{71.93} & \textcolor{PineGreen}{94.25} & \textcolor{PineGreen}{79.16} & \textcolor{PineGreen}{75.89} & \textcolor{PineGreen}{75.27} & \textcolor{PineGreen}{89.82} & 68.17 & \textcolor{red}{63.54} & \textcolor{red}{62.94} & \textcolor{PineGreen}{78.77} \\
 & L2 & 89.68 & \textbf{\textcolor{PineGreen}{91.99}} & \textcolor{red}{83.71} & \textcolor{PineGreen}{94.25} & \textcolor{PineGreen}{90.11} & \textcolor{PineGreen}{77.85} & \textcolor{red}{71.70} & \textbf{\textcolor{PineGreen}{94.40}} & \textcolor{PineGreen}{79.20} & \textcolor{PineGreen}{75.89} & \textbf{\textcolor{PineGreen}{75.32}} & \textcolor{PineGreen}{90.14} & 66.98 & 66.88 & \textcolor{red}{62.22} & \textcolor{PineGreen}{78.75} \\
 & Dropout & 90.60 & \textcolor{red}{90.24} & 84.92 & \textcolor{PineGreen}{94.75} & \textcolor{PineGreen}{89.61} & \textbf{\textcolor{PineGreen}{85.78}} & \textcolor{red}{71.89} & \textcolor{PineGreen}{93.27} & \textcolor{PineGreen}{79.23} & 74.13 & \textcolor{red}{72.31} & \textcolor{PineGreen}{91.01} & \textcolor{PineGreen}{68.64} & \textcolor{red}{63.36} & \textcolor{red}{66.10} & \textbf{\textcolor{PineGreen}{79.17}} \\
 & LP & \textbf{92.20} & 91.28 & \textcolor{red}{83.97} & \textcolor{PineGreen}{78.89} & \textcolor{PineGreen}{74.23} & 72.94 & \textcolor{red}{72.15} & \textcolor{PineGreen}{75.18} & 74.85 & 74.69 & 74.72 & \textcolor{PineGreen}{71.71} & 67.69 & 67.67 & \textcolor{red}{67.20} & \textcolor{PineGreen}{72.66} \\
 & LP-FT & 90.37 & \textcolor{PineGreen}{91.69} & \textcolor{red}{83.69} & \textcolor{PineGreen}{93.98} & \textcolor{PineGreen}{88.93} & \textcolor{PineGreen}{76.23} & \textcolor{red}{71.57} & \textcolor{PineGreen}{93.80} & \textcolor{PineGreen}{78.33} & \textbf{\textcolor{PineGreen}{76.89}} & 74.84 & \textcolor{PineGreen}{90.20} & 67.71 & 67.61 & \textcolor{red}{62.00} & \textcolor{PineGreen}{78.51} \\
 & IRM & 90.37 & \textcolor{red}{90.17} & \textcolor{red}{82.21} & \textcolor{PineGreen}{94.93} & \textcolor{PineGreen}{88.86} & \textcolor{PineGreen}{81.81} & 72.09 & \textcolor{PineGreen}{93.74} & \textcolor{PineGreen}{79.88} & \textcolor{PineGreen}{75.64} & \textcolor{red}{73.57} & \textcolor{PineGreen}{89.54} & \textcolor{PineGreen}{69.86} & \textcolor{red}{66.39} & \textcolor{red}{29.84} & \textcolor{PineGreen}{76.34} \\
 & DRO & \textcolor{red}{88.53} & \textcolor{red}{88.37} & \textcolor{red}{78.43} & \textcolor{PineGreen}{93.92} & \textcolor{PineGreen}{89.40} & \textcolor{PineGreen}{81.51} & \textcolor{red}{69.50} & \textcolor{PineGreen}{92.87} & \textcolor{PineGreen}{76.97} & \textcolor{red}{73.75} & \textcolor{red}{72.72} & \textcolor{PineGreen}{86.61} & \textcolor{red}{64.02} & \textcolor{red}{61.42} & \textcolor{red}{59.66} & \textcolor{PineGreen}{76.86} \\
 & Fish & 89.91 & \textcolor{red}{90.74} & \textcolor{red}{82.46} & \textcolor{PineGreen}{94.69} & \textbf{\textcolor{PineGreen}{91.39}} & \textcolor{PineGreen}{76.19} & \textcolor{red}{71.84} & \textcolor{PineGreen}{94.19} & \textcolor{PineGreen}{78.94} & \textcolor{PineGreen}{76.35} & 74.20 & \textcolor{PineGreen}{89.52} & \textcolor{PineGreen}{69.19} & 67.70 & \textcolor{red}{62.10} & \textcolor{PineGreen}{78.86} \\
    \bottomrule
   \end{tabularx}
   \caption{I.i.d.\ test set performance and generalisation measures (in \%) of each examined method for all tasks and training configurations. The Avg. column shows the average \g score across all tasks and generalisation measures. We show scores significantly above and below the IID baseline (first row, suite scores are \gstandard) in green and red, respectively, and write the best score for each column in bold weight. When the score is not significantly different from the baseline counterpart, we show it in black. We use two-tailed binomial testing when comparing the i.i.d.\ performances, and randomisation testing \cite{yeh2000accurate} when comparing G scores, setting $0.05$ as the significance level.
   }\label{tab:aggResults}
\end{table*}

\begin{figure*}[tb]
  \centering
  \includegraphics[width=\textwidth]{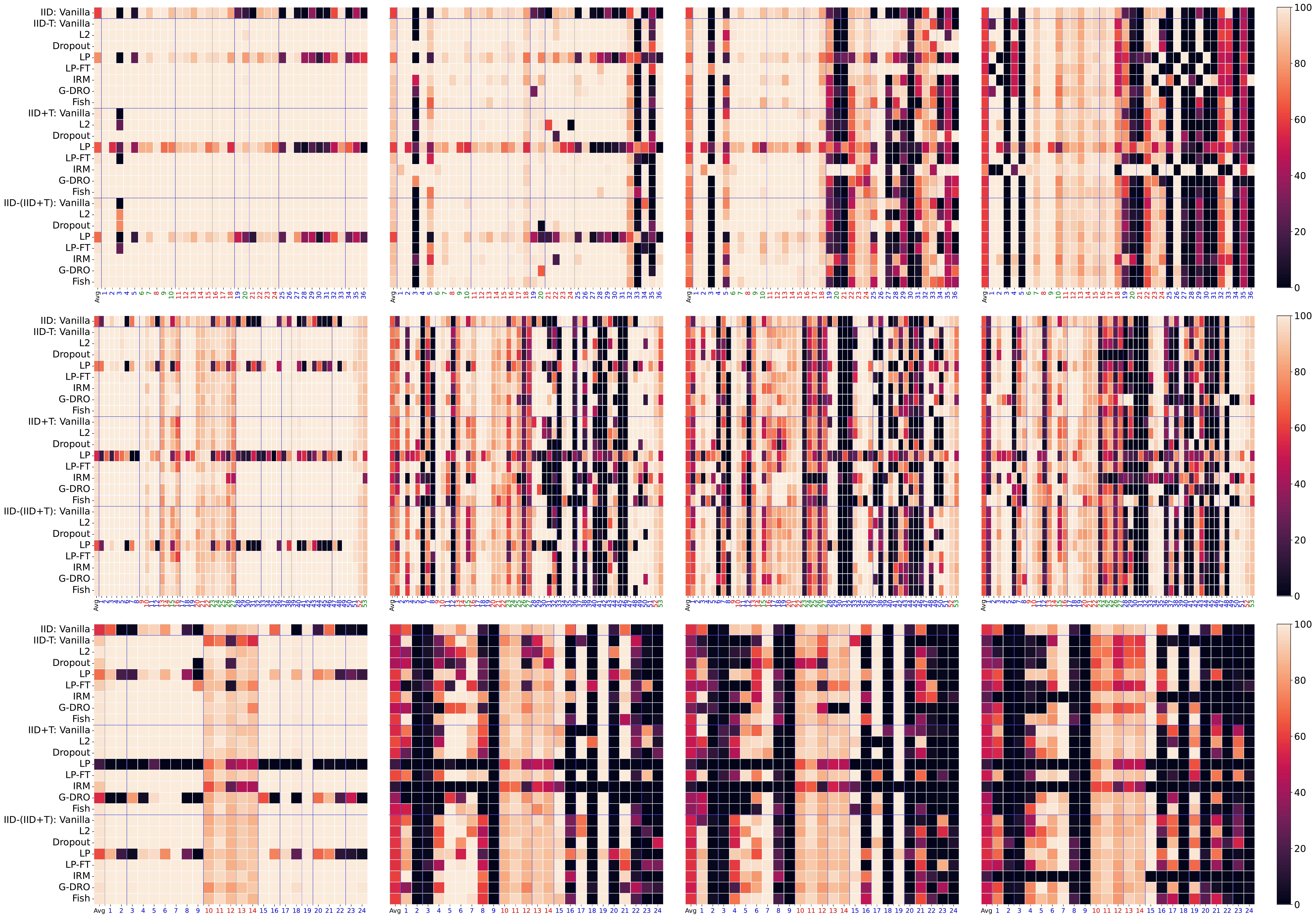}
     \caption{Average and individual pass rates for all tasks, methods and training configurations. From first to third row: results for \texttt{SENT}, \texttt{PARA} and \texttt{READ}. From first to fourth column: seen evaluation, functionality generalisation, functionality class generalisation, and test type generalisation scores. The y-axis correspond to all training configuration-method pairs; the x-axis shows the average functionality pass rate followed by the individual pass rates. The blue horizontal and vertical lines demarcate different training configurations and functionality classes, respectively. The colors in the x-axis designate the different test types: blue for MFTs, red for INVs an green for DIRs.}
     \label{fig:fineResults}
\end{figure*}

\subsection{I.i.d.\ and generalisation scores}
Table~\ref{tab:aggResults} exhibits i.i.d.\ and aggregate \g scores for all tasks, training configurations and generalisation methods. Figure~\ref{fig:fineResults} presents pass rates of individual functionalities.

% Seen evaluation
\textbf{Seen performance}:~
Fine-tuning on test suite data led to improvements for all tasks: the \gseen scores are generally higher than the baseline scores (first row in Table~\ref{tab:aggResults}).

That is, models were able to generalise across test cases from covered functionalities (from  \Ttrain to \Ttest) while retaining reasonable i.i.d.\ data performance.
In some specific training configuration-method combinations this was not the case. We discuss this below when we compare methods and report the degenerate solutions.

% Q2 results
\textbf{Generalisation performance}:~
% Q2 (generalisation Analysis). Seen score is consistently over-optimistic (when compared to the generalisation metrics). Consistent less generalisation from func to class to aspect.
For any given configuration-method pair, \gseen is higher than \gfunc, \gclass and \gtype, indicating a generalisation gap between seen and unseen functionalities.
Furthermore, for all tasks, average (across methods) \gfunc is higher than average \gclass, which is higher than average \gtype,\footnote{\texttt{SENT}: 85.97/78.15/69.54, \texttt{PARA}: 75.04/72.22/71.55, \texttt{READ}: 49.23/46.66/43.46.} indicating that generalisation gets harder as one moves from unseen functionalities to unseen functionality classes and test types.
This aligns with previous work \cite{luzdearaujo2022checking}, in which hate speech detection models are found to generalise within---but not across---functionality classes.

Improvements over the IID baseline were task dependent.
Almost all configuration-method pairs achieved \gfunc (22 of 24) and \gclass (20 of 24) scores significantly higher that the IID baseline for \texttt{SENT}, with improvements over the baseline as high as 18.44 and 12.84 percentage points (p.p.) for each metric, respectively.
For \texttt{PARA}, improving over \gclass proved much harder---only seven configuration-method pairs could do so.
Increases in score were also less pronounced, the best \gfunc and \gclass scores being 6.91 and 2.19 p.p.\ above the baseline.
\texttt{READ} was the one with both rarer and subtler improvements, with a third of the approaches significantly improving functionality and none significantly improving functionality class generalisation.
Improvements in  each case were as high as 4.70 and 0.51 p.p.\ over the baseline.

%Q3 results
\textbf{I.i.d.\ performance}:~
Fine-tuning on test suite data only (\iidAndT configuration) reduced performance for all tasks' i.i.d.\ test sets.
Fine-tuning on both suite and i.i.d.\ examples (\iidPlusT and \iidAndIidPlusT) helped retain---or improve---performance in some cases, but decreases were still more common.
The \iidAndIidPlusT configuration was the most robust regarding i.i.d.\ scores, with an average change (compared to the IID baseline) of $-1.43$/$-0.50$/$-1.73$ for \texttt{SENT}/\texttt{PARA}/\texttt{READ}.  

\subsection{Training configuration and method comparison}
\label{sec:methodDisc}
Using a mixture of i.i.d.\ and suite samples proved essential to retain i.i.d.\ performance: the overall scores (average over methods and i.i.d.\ test sets) for each configuration are 67.52, 76.33 and 87.98 for \iidAndT, \iidPlusT, and \iidAndIidPlusT respectively.

That said, the environment-based generalisation algorithms (IRM, DRO and Fish) struggled in the \iidPlusT configuration, underperforming when compared with the other methods.
We hypothesize that in these scenarios models simply do not see enough i.i.d.\ data, as we treat it as just one more environment among many others (reaching as much as 54 in \texttt{PARA}).
LP also achieves subpar scores, even though i.i.d.\ data is not undersampled.
The problem here is the frozen feature encoder, as BERT features are not good enough without fine-tuning on i.i.d.\ task data---as was done in the other configurations, with clear benefits for LP.

No individual method performed best for all scores and tasks.
That said, \iidAndIidPlusT with L2, LP, LP-FT or Fish was able to achieve \gfunc and \gclass scores higher or not significantly different from the baseline in all tasks, though \iidAndIidPlusT with dropout was the best when score is averaged over all tasks and generalisation measures.
Considering this same metric, \iidAndIidPlusT was the most consistently good configuration, with all methods improving over the average IID baseline.

% \subsection{Qualitative analysis}
% \label{sec:qual}

% \textbf{Stubborn and volatile functionalities}
% % Insights from image: Functionalities that are consistently hard to generalise and those that the model got right and then got wrong. Maybe one case example from each task.
% \added{
  
% }

\subsection{DIR applicability}
We have found that DIRs, as used for \texttt{SENT}, have limited applicability for both testing and training.
The reason for that is that models are generally very confident about their predictions: the average prediction confidence for the test suite predictions is $0.97$ for the IID model.
On the evaluation side, this makes some DIRs impossible to fail: the confidence cannot get higher and fail ``not more confident'' expectations.
On the training side, DIRs do not add much of a training signal, as the training loss is near zero from the very beginning.\footnote{Confidence regularisation \cite{yu2021fine} could potentially increase DIR's usefulness for training and evaluation purposes.}

We see an additional problem with DIRs in the \texttt{SENT} setting: they confuse prediction confidence with sentiment intensity.
% On the evaluation side, they confuse prediction confidence with sentiment intensity.
Though prediction confidence may correlate with sentiment intensity, uncertainty also signals difficulty and ambiguousness \cite{swayamdipta2020dataset}.
Consequently, sentiment intensity tests may not be measuring the intended phenomena.
One alternative would be to disentangle the two factors: using prediction values only for confidence-based tests, and sentiment intensity tests only for sentiment analysis tasks with numeric or fine-grained labels.

\subsection{Negative transfer}

Though \gclass scores are generally lower than \gfunc scores, this is not always the case for the pass rates of individual functionalities.
When there are contrastive functionalities within a class---those whose test cases have similar surface form but entirely different expected behaviours---it is very difficult to generalise from one to the other.

For example, the SRL class in \texttt{PARA} contains the functionalities ``order does not matter for symmetric relations'' and ``order does matter for asymmetric relations'' (functionalities 41 and 42 in the second row of Fig.~\ref{fig:fineResults}).
Their test cases are generated by nearly identical templates where the only change is the relation placeholder. 
Examples from the first and second functionalities would include (Q1: Is Natalie dating Sophia? Q2: Is Sophia dating Natalie?) and (Q1: Is Matthew lying to Nicole? Q2: Is Nicole lying to Matthew?) respectively.
Though their surface forms are similar, they have opposite labels: duplicate and not duplicate.

To compute $s_{\T\text{func}}$, a model is trained with samples from one functionality and evaluated on samples from the other.
Consequently, the surface form will be spuriously correlated with the label seen during training and models may blindly assign it to the question pairs that fit the template.
This would work well for the seen functionality, but samples from the unseen one would be entirely misclassified.
Conversely, when computing the $s_{\T\text{class}}$ score, the model will not have been trained on either of the functionalities and will not have the chance to adopt the heuristic, leading to better unseen pass rates.

\subsection{Degenerate solutions}
Settings where the \gtype score is higher than the baseline are much rarer than for the other measures, happening only in one case for \texttt{SENT} (\iidAndT with dropout) and never for \texttt{READ}.
One explanation is that training only on perturbation-based tests (with no MFTs) can lead to degenerate solutions, such as passing all tests by always predicting the same class.

To assess if that was the case, we examined the predictions on the SST-2 test set of the \iidAndT vanilla model fine-tuned only on DIRs and INVs.
We have found that $95.18\%$ of the i.i.d.\ data points were predicted as negative, though the ground truth frequency for that label is $47.25\%$.
When examining the predictions for MFTs, the results are even more contrasting: $0.29\%$ of the predictions were negative, with the ground truth frequency being $43.42\%$.
These results show that the model has, indeed, adopted the degenerate solution.
Interestingly, it predicts different classes depending on the domain, almost always predicting negative for i.i.d.\ data and positive for suite data.

The gap between \gclass and \gtype scores in \texttt{PARA} is not as severe, possibly due to the supervised signal in its DIRs.
Since these tests expect inputs to correspond to specific labels---as opposed to DIRs for \texttt{SENT}, which check for changes in prediction confidence---always predicting the same class would not be a good solution.
Indeed, when examining the predictions on the QQP test set of the vanilla \iidAndT model fine-tuned with no MFT data, we see that $58.70\%$ of question pairs are predicted as not duplicate, which is similar to the ground truth frequency, $63.25\%$.
The same is true when checking the predictions for MFTs: $64.47\%$ of the data points are predicted as not duplicate, against a ground truth frequency of $52.46\%$.

The \texttt{READ} scenario is more complex---instead of categories, spans are extracted.
Manual inspection showed that some \iidAndT models adopted degenerate solutions (e.g.\  extracting the first word, a full stop or the empty span as the answer), even when constrained by the MFT supervised signal.
Interestingly, the degenerate solutions were applied only for INV tests (where such invariant predictions work reasonably) and i.i.d.\ examples (where they do not).
On the other hand, these models were able to handle the MFTs well, obtaining near perfect scores and achieving high $s_{\T\text{seen}}$ scores even though i.i.d.\ performance is catastrophic.
The first grid of the third row in Fig.~\ref{fig:fineResults} illustrates this: the high $s_{\T\text{seen}}$ scores are shown on the first column, and the MFT pass rates on the columns with blue x-axis numbers.
% Using \gseen as the metric for comparison accounts for that, yielding much lower scores, as shown by the values corresponding to the \iidAndT configuration in the \gseen column from the \texttt{READ} task in Table~\ref{tab:aggResults}.}

\subsection{Summary interpretation of the results}
\paragraph{Figure~\ref{fig:fineResults}}
Figure~\ref{fig:fineResults} supports fine-grained analyses that consider performance on individual functionalities in each generalisation scenario.
One can interpret it horizontally to assess the functionality pass rates for a particular method. For example, the bottom left grid, representing seen results for \texttt{READ}, shows that \iidPlusT with LP behaves poorly on almost all functionalities, confirming the importance of fine-tuning BERT pre-trained features  (§ \ref{sec:methodDisc}).

Alternatively, one can interpret it vertically to assess performance and generalisation trends for individual functionalities.
For example, models generalised well to functionality 21 of the \texttt{READ} suite (second grid of the bottom row), with most methods improving over the IID baseline.
However, under the functionality class evaluation scenario (third grid of the bottom row), improvements for functionality 21 are much rarer.
That is, the models were able to generalise to functionality 21 as long as they were fine-tuned on cases from functionalities from the same class (20 and 22)\footnote{These functionalities assess co-reference resolution capabilities: 20 and 21 have test cases with personal and possessive pronouns, respectively; 22 tests whether the model distinguishes ``former'' from ``latter''.}.

Such fine-grained analyses show the way for more targeted explorations of generalisation (e.g.\ why do models generalise to functionality 21 but not to functionality 20?), which can guide subsequent data annotation, selection and creation efforts, and shed light on model limitations.

% Func 20:	Basic coref, he / she
    % C: Antonio and Riley are friends. She is a photographer, and he is an attorney.
    % Q: Who is an attorney?
    % A: Antonio
% Func 21: Basic coref, his / her
    % C: Caleb and Kathryn are friends. His mom is an economist.
    % Q: Whose mom is an economist?
    % A: Caleb

\paragraph{Table~\ref{tab:aggResults}}
For i.i.d.\ results, we refer to the SST2, QQP and SQuAD columns. These show that the suite-augmented configuration and methods (all rows below and including \iidAndT Vanilla) generally hurt i.i.d.\ performance. 
However, improvements can be found for some methods in the \iidPlusT and \iidAndIidPlusT.
\textbf{Takeaway: fine-tuning on behavioural tests degrades model general performance, which can be mitigated by jointly fine-tuning on i.i.d.\ samples and behavioural tests.}

For performance concerning seen functionalities, we refer to the \gseen columns. 
Generalisation scores concerning unseen functionalities, functionality classes and test types can be found in the \gfunc, \gclass and \gtype columns. 
Across all tasks, training configurations and methods, the \gseen scores are higher than the others.
\textbf{Takeaway: evaluating only on the seen functionalities \cite{liu_etal_InocFineTune_NACACL_2019,malon2022fast} is overoptimistic---improving performance on seen cases may come at the expense of degradation on unseen cases. This is detected by the underperforming generalisation scores.}

Previous work on generalisation in behavioural learning \cite{luzdearaujo2022checking,rozen2019diversify} corresponds to the \iidAndT Vanilla row.
It shows deterioration of i.i.d.\ scores, poor generalisation in some cases, and lower average performance compared with the IID baseline.
However, our experiments with additional methods (all rows below \iidAndT Vanilla), show that some configuration-method combinations improve the average performance.
\textbf{Takeaway: while naive behavioural learning generalises poorly, more sophisticated algorithms can lead to improvements. \beluga is a method that detects and measures further algorithmic improvements.}

\section{Related work}

% Papers that enables approach
% Behavioural tests.
Traditional NLP benchmarks \cite{wang2018glue, wang2019superglue} are composed of text corpora that reflect the naturally-occurring language distribution, which may fail to sufficiently capture rarer, but important phenomena \cite{belinkov_glass_analNLP_TACL_2019}.
Moreover, since these benchmarks are commonly split into identically distributed train and test sets, spurious correlations in the former will generally hold for the latter. 
This may lead to the obfuscation of unintended behaviours, such as the adoption of heuristics that work well for the data distribution but not in general \cite{linzen_accelerateHLLG_ACL_2020,mccoy_etal_right4wrongreasons_ACL_20219}.
To account for these shortcomings, complementary evaluations methods have been proposed, such as using dynamic benchmarks \cite{kiela2021dynabench} and behavioural test suites \cite{kirk2022hatemoji,rottger_etal_hateCheck_ACL_2021,ribeiro_etal_checklist_ACL_2020}.

A line of work has explored how training on challenge and test suite data affects model performance by fine-tuning on examples from specific linguistic phenomena and evaluating on other samples from the same phenomena \cite{malon2022fast,liu_etal_InocFineTune_NACACL_2019}. This is equivalent to our seen evaluation scenario, and thus cannot distinguish between models with good generalisation and those that have overfitted to the seen phenomena. We account for that with our additional generalisation measures, computed using only data from held-out phenomena.

% \mbox{\citet{liu_etal_InocFineTune_NACACL_2019}} have proposed inoculation by fine-tuning, where model and dataset shortcomings are examined by fine-tuning the model on challenge data samples corresponding to specific phenomena (e.g., negation) and comparing the performance before and after the procedure on both challenge and non-challenge datasets.
% \citet{malon2022fast} have created a method for using suite data to increase accuracy on covered phenomena, reporting substantial performance gains.
% Both methods, however, train and evaluate on samples assigned to the same phenomena and therefore cannot detect models that have overfitted to the test cases.

Other efforts have also used controlled data splits to examine generalisation: \citet{mccoy_etal_right4wrongreasons_ACL_20219} have trained and evaluated on data from disjoints sets of phenomena relevant for Natural Language Inference (NLI); \citet{rozen2019diversify} have split challenge data according to sentence length and constituency parsing tree depth, creating a distribution shift between training and evaluation data; \citet{luzdearaujo2022checking} employ a cross-functional analysis of generalisation in hate speech detection.
Though these works address the issue of overfitting to seen phenomena, their analyses are restricted to specific tasks and training configurations.
Our work gives a more comprehensive view of generalisation of behavioural learning by examining different tasks, training configurations, test types and metrics.
Additionally, we use this setting as an opportunity to compare generalisation impact of both simple regularisation mechanisms and state-of-the-art domain generalisation algorithms.

% \replaced{The idea of cross-functional analysis of behavioural learning has been originally proposed by \mbox{\citet{luzdearaujo2022checking}}, where generalisation is measured by training and evaluating models on controlled data splits that hold out sets of suite functionalities for evaluation.
% While they examine several dimensions of generalisation (across functionalities, functionality classes and protected groups), their analysis was restricted to the hate speech detection task, the fine-tuning training configuration and to one test type.
% Furthermore, their analysis does not include fine-grained results: only overall accuracy scores are reported, which obfuscates performance on less frequent phenomena.

% In this work, we are the first to comprehensively examine generalisation of behavioural learning in different tasks, training configurations, test types and axes of generalisation.
% Contrary to previous work, we also compare the impact of several regularisation and generalisation algorithms on unseen phenomena performance.}

\section{Conclusion}
We have presented \beluga, a framework for cross-functional analysis of generalisation in NLP systems that both makes explicit the desired system traits and allows for quantifying and examining several axes of generalisation.
While in this work we have used \beluga to analyse data from behavioural suites, it can be applied in any setting where one has access to data structured into meaningful groups (e.g.\ demographic data, linguistic phenomena, domains).
% We have fine-tuned and evaluated pre-trained models on several partitions of test suites to estimate generalisation to unseen functionalities and functionality classes.

We have shown that, while model performance for seen phenomena greatly improves after fine-tuning on test suite data, the generalisation scores reveal a more nuanced view, in which the actual benefit is less pronounced and depends on the task and training configuration-method combination.
We have found the \iidAndIidPlusT configuration to result in the most consistent improvements. Conversely, some methods struggle in the \iidAndT and \iidPlusT settings by overfitting to the suite or underfitting i.i.d.\ data, respectively.
In these cases, a model both practically aces all tests and fails badly for i.i.d.\ data, which reinforces the importance of considering both i.i.d.\ and test suite performance when comparing systems, which is accounted for by \beluga's aggregate scores.

These results show that naive behavioural learning has unintended consequences, which the \iidAndIidPlusT configuration mitigates to some degree.
There is still much room for improvement, though, especially if generalisation to unseen types of behaviour is desired.
Through \beluga, progress in that direction is measurable, and further algorithmic improvements might make behavioural learning an option to ensure desirable behaviours and preserve general performance and generalisability of the resulting models.
We do not recommend training on behavioural tests in the current technological state. Instead, we show a way to improve research on reconciling the qualitative guidance of behavioural tests with desired generalisation in NLP models.

\section*{Acknowledgements}
We thank the anonymous reviewers and action editors for the helpful suggestions and detailed comments.
We also thank Matthias Aßenmacher, Luisa März, Anastasiia Sedova, Andreas Stephan, Lukas Thoma, Yuxi Xia, and Lena Zellinger for the valuable discussions and feedback. This research has been funded by the Vienna Science and Technology Fund (WWTF) [10.47379/VRG19008] ``Knowledge-infused Deep Learning for Natural Language Processing''.

\bibliography{main}
\bibliographystyle{acl_natbib}

\end{document}